\documentclass[11pt]{article}

\usepackage[preprint]{acl}

\usepackage{times}
\usepackage{latexsym}
\usepackage{amsmath}
\usepackage{amsfonts}

\usepackage[T1]{fontenc}

\usepackage[utf8]{inputenc}

\usepackage{microtype}

\usepackage{inconsolata}

\usepackage{graphicx}

\usepackage{hyperref}
\usepackage{url}
\usepackage{tabularx}
\usepackage{booktabs}
\usepackage{wrapfig} 
\usepackage{multirow}
\usepackage{caption}
\usepackage{enumitem}
\usepackage[most]{tcolorbox}

\title{\includegraphics[height=1.4em]{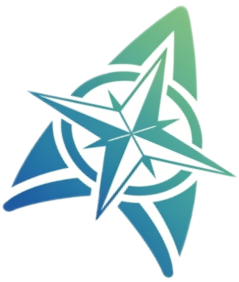} MapTrace: Scalable Data Generation for Route Tracing on Maps}

\author{Artemis Panagopoulou$^\nabla$\thanks{\tiny Work done during internship at Google.} \qquad
  Aveek Purohit$^\Box$ \qquad Achin Kulshrestha$^\Box$ \\\textbf{Soroosh Yazdani}$^\Box$ \qquad \textbf{Mohit Goyal}$^\Box$\\
   $\Box$ Google XR \qquad $\nabla$ University of Pennsylvania\\
  \small{Project Page:   \href{https://artemisp.github.io/maptrace}{https://artemisp.github.io/maptrace}}}

\begin{document}
\maketitle
\begin{abstract}
While Multimodal Large Language Models have achieved human-like performance on many visual and textual reasoning tasks, their proficiency in fine-grained spatial understanding, such as route tracing on maps remains limited. Unlike humans, who can quickly learn to parse and navigate maps, current models often fail to respect fundamental path constraints, in part due to the prohibitive cost and difficulty of collecting large-scale, pixel-accurate path annotations. To address this, we introduce a scalable synthetic data generation pipeline that leverages synthetic map images and pixel-level parsing to automatically produce precise annotations for this challenging task. Using this pipeline, we construct a fine-tuning dataset of 23k path samples across 4k maps, enabling models to acquire more human-like spatial capabilities. Using this dataset, we fine-tune both open-source and proprietary MLLMs. Results on MapBench show that finetuning substantially improves robustness, raising success rates by up to 6.4 points, while also reducing path-tracing error (NDTW). These gains highlight that fine-grained spatial reasoning, absent in pretrained models, can be explicitly taught with synthetic supervision.
\end{abstract}

\section{Introduction}
\vspace{-.2cm}
Multimodal Large Language Models (MLLMs) have achieved impressive progress in joint visual and textual understanding, demonstrating strong capabilities across recognition, reasoning, and instruction following tasks~\citep{team2023gemini,li2022blip,bai2025qwen2,team2025gemma}. Despite these advances, their spatial reasoning ability remains underdeveloped. While models excel at semantic tasks such as identifying objects and describing visual scenes, they perform poorly on tasks that require reasoning about spatial, geometric, and topological concepts~\citep{kamath2023s,ro2025well}.
\begin{figure}[ht]
\centering \includegraphics[width=.9\linewidth]{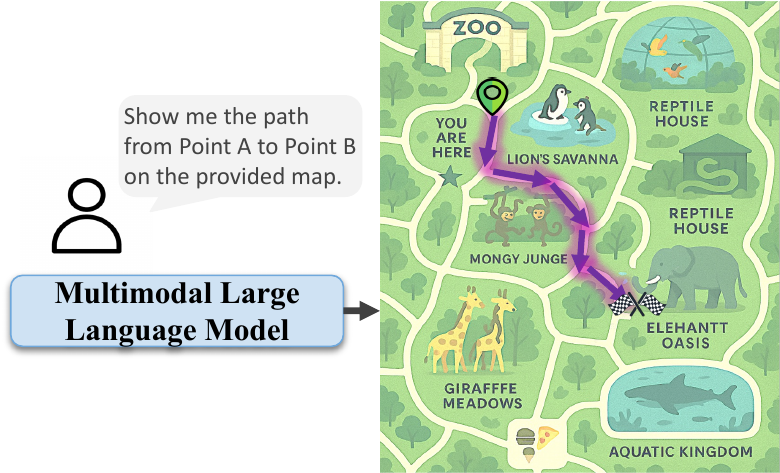}
\vspace{-0.2cm}
\caption{MapTrace: Given a start and end location, the model outputs a valid path that respects map constraints}
\vspace{-.6cm}
\label{fig:pathfinding_sample}
\end{figure}
In contrast to humans, who quickly learn to parse and use maps, MLLMs often fail to produce valid routes, revealing a persistent limitation in their current training paradigms.
To address this deficiency, we introduce MapTrace, a new task that requires a model to generate a precise sequence of coordinates forming a valid path between two points on a map (Figure~\ref{fig:pathfinding_sample}). Unlike high-level navigation tasks that rely on language instructions~\citep{xing2025can,mukhopadhyay2025mapwise}, MapTrace isolates a model’s ability to reason directly about connectivity and traversability at the pixel level, grounding model's understanding of maps.
A major challenge towards effectively teaching MLLMs to trace routes on maps  is the prohibitive cost of creating large-scale, pixel-accurate path annotations and the difficulty in obtaining usage permission of layout maps from commercial sources. To overcome this, we design a scalable synthetic data generation pipeline. The pipeline generates diverse map layouts, automatically constructs traversable path masks, converts them into graph structures, and produces optimal shortest-path annotations. This process yields 23k high-quality path samples across 4k maps, providing explicit supervision for connectivity, traversability, and pathfinding. Fine-tuning MLLMs on this dataset results in up to a 30\% relative improvement on MapBench~\citep{xing2025can}.
 Our contributions are threefold:\\
(1) We introduce \textbf{MapTrace}, the first task and dataset explicitly designed to probe coordinate-level spatial reasoning in MLLMs by requiring models to generate pixel-accurate paths on maps. \\ 
(2) We present a scalable \textbf{synthetic data generation pipeline} that produces 23k path annotations over 4k diverse maps, leveraging LLM-based critics to ensure high fidelity.  \\
(3) We provide \textbf{empirical insight} into the spatial reasoning gap in current MLLMs, showing that fine-tuning on MapTrace substantially improves robustness (raising success rates by up to 6.4 points) and reduces path-tracing error (NDTW) in most map domains.
\vspace{-.2cm}
\section{Related Work}
\vspace{-.23cm}
\noindent\textbf{Spatial Reasoning.} MLLMs have achieved strong results across recognition, reasoning, and instruction-following tasks~\citep{team2023gemini,li2022blip,bai2025qwen2,team2025gemma}, yet many studies highlight persistent weaknesses in spatial reasoning. Benchmarks targeting geometric relations, relative positioning, and layout understanding~\citep{wang2024picture,yamadaevaluating,fuisobench} consistently show that models falter when asked to reason about structure rather than appearance. For example, even when models can identify objects correctly, they struggle to judge whether one object is inside, adjacent, or aligned with another. These limitations suggest that current pre-training provides limited supervision for spatial structure~\citep{kamath2023s}. Our work focuses on one fundamental instance of this challenge, path-tracing on maps, serving as a probe of fine-grained spatial reasoning.

\noindent\textbf{Map-Based Reasoning.}
Maps provide a natural testbed for spatial reasoning because they require both visual parsing and structured interpretation. Prior work has explored constrained domains such as floor plans with explicit labels~\citep{defazio2024vision,kalervo2019cubicasa5k}, standardized geographic maps~\citep{zhang2025mapreader,mukhopadhyay2025mapwise}, and transit networks~\citep{fang2024travellm}. These tasks take the form of either question answering about landmarks, or generating high-level navigation instructions~\citep{li2024semantic,fang2024travellm}, with a smaller body of work attempting explicit path generation but only under simplified layouts~\citep{defazio2024vision,kalervo2019cubicasa5k}. More complex commercial maps (e.g., malls, zoos, museums) have received less attention, with exceptions such as \citet{coffrinillm}, who evaluated MLLMs on three maps and found frequent failures in spatial grounding. MapBench~\citep{xing2025can} introduced a broader benchmark on commercial maps, measuring sequences of correct landmarks as a proxy for navigation. Our work complements this line by shifting evaluation to the coordinate level, where path quality is judged by geometric accuracy and adherence to map constraints.

\noindent\textbf{Synthetic Data Generation.}
The scarcity of large-scale, pixel-accurate path annotations poses a major challenge for training models on map reasoning. To address similar bottlenecks in other domains, researchers have increasingly turned to synthetic data generation, where LLMs and MLLMs act both as data producers~\citep{liu2024synthvlm,chen2024allava} and evaluators or ``judges”~\citep{long2024llms}. This paradigm has been used to scale instruction-following datasets~\citep{li2024synthetic}, align models with human preferences~\citep{dong2025selfboosting}, and filter low-quality samples~\citep{jin2024optimizing}. Synthetic data has also proven valuable for tasks that are expensive or ambiguous to annotate, such as compositional reasoning~\citep{mishra2025scramble}. We extend these ideas to the spatial domain: our pipeline generates diverse synthetic maps, derives shortest-path annotations automatically, and uses MLLM critics for filtering low quality samples. This pipeline allows us to efficiently produce high-quality data for path-tracing, a capability that has been prohibitively expensive to collect data at scale due to licencing restrictions and high annotation costs. 
\vspace{-.2cm}
\section{Novel Task: MapTrace}
\vspace{-.2cm}

We introduce \textbf{MapTrace}, a task designed to evaluate the fine-grained spatial reasoning abilities of MLLMs. Unlike prior work that focuses on simplified layouts~\citep{kalervo2019cubicasa5k} or high-level instruction following~\citep{xing2025can}, MapTrace requires models to generate \textbf{pixel-accurate, traversable paths} on commercial wayfinding maps (e.g., malls, zoos, museums). Figure~\ref{fig:pathfinding_sample} illustrates the setup: given start and end locations, the model must output a valid route that respects the map’s connectivity constraints.

\begin{figure*}[t]
    \centering
    \vspace{-.1cm}
    \includegraphics[width=\linewidth]{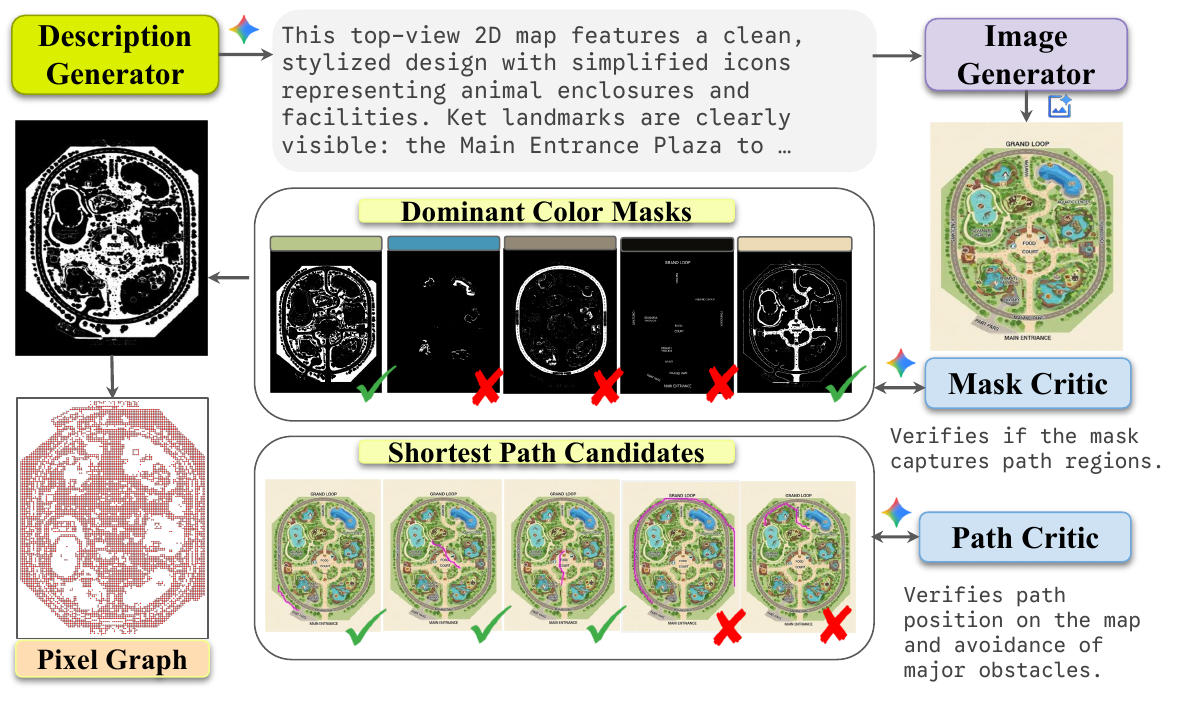}
    \vspace{-0.8cm}\caption{MapTrace Synthetic data generation pipeline. A LLM generates various map descriptions, which are rendered into images by a text-to-image model. Candidate path masks are extracted via dominant-color selection and filtered by a Mask Critic. Valid masks are converted into a pixel-graph to compute shortest-path candidates, which are then judged by a Path Critic for quality and traversability.}
    \vspace{-0.4cm}
    \label{fig:data_generation}
\end{figure*}

\vspace{-.2cm}
\subsection{Task Definition}
\vspace{-.1cm}
Let $\mathcal{I} \in \mathbb{R}^{H \times W \times 3}$ denote an RGB wayfinding map of width $W$ and height $H$. Each map is associated with a binary traversability mask $\mathcal{M} \in \{0,1\}^{H \times W}$, where $\mathcal{M}(x,y) = 1$ if pixel $(x,y)$ is traversable and $\mathcal{M}(x,y) = 0$ otherwise. The task input consists of a start coordinate $s = (x_s,y_s)$ and an end coordinate $t = (x_t,y_t)$. The model must output an \textbf{ordered sequence of coordinates} 
$P = \{(x_1,y_1), (x_2,y_2), \dots, (x_k,y_k)\}$,
such that: (1) $P$ forms a continuous path from $s$ to $t$, (2) every $(x_i,y_i) \in P$ satisfies $\mathcal{M}(x_i,y_i)=1$ (i.e., lies within traversable regions), and (3) $P$ does not intersect obstacles or leave the navigable area.

\noindent\textbf{Ground truth paths} are defined as shortest routes between $s$ and $t$ under $\mathcal{M}$, represented as sequences of pixel coordinates that trace traversable areas. Humans can easily complete this task by visually following map pathways/corridors, but current MLLMs perform poorly, highlighting a key gap in map reasoning ability.
\vspace{-.2cm}
\subsection{Evaluation Metrics}
\vspace{-.1cm}
\noindent\textbf{Normalized Dynamic Time Warping (NDTW).}  
We evaluate path quality using a normalized variant of Dynamic Time Warping (DTW)~\citep{magalhaes2019general}, which flexibly aligns predicted and ground-truth coordinate sequences even if they differ in length and the images differ in resolution. To ensure scale invariance across maps, all coordinates are normalized to $[0,1]$ by dividing by map width and height. The resulting NDTW distance measures geometric similarity between the predicted and true paths, with lower values indicating better alignment.  

\noindent\textbf{Success Rate (SR).}  
We additionally report the percentage of valid paths, where a path is considered unsuccessful if no coordinates are produced, the output cannot be parsed, or NDTW is undefined.

\begin{table*}[t]
\centering
\resizebox{\textwidth}{!}{ 
\begin{tabular}{lccccccccc|cc}
\toprule
\textbf{Method} & \textbf{Google} & \textbf{Mall} & \textbf{Museum} & \textbf{Nat. Park} & \textbf{Theme} & \textbf{Trail} & \textbf{Campus} & \textbf{Urban} & \textbf{Zoo} & \textbf{Avg.} & \textbf{SR (\%)} \\
\midrule
\multicolumn{12}{l}{\textit{Baseline}} \\
Mask-Method & 0.91 & 0.75 & 0.67 & 0.91 & 0.68 & 0.65 & 0.64 & 0.78 & 0.90 & 0.79 & 73.4 \\
\midrule
\multicolumn{12}{l}{\textit{Pretrained Models}} \\
Gemini 2.5 Pro & 1.25 & 0.87 & 1.16 & 1.18 & 1.57 & 1.06 & 1.13 & 1.34 & 0.80 & 1.19 & 90.6 \\
 GLM4.5V & 2.34&10.79&3.06&0.71&2.42&1.83&0.51&5.20&1.25&2.30&70.7\\
QwenVL-2.5 72B &2.12&0.73& 0.90 & 1.01 & 1.19 & 1.34 & 1.05& 0.81 &1.71  &1.32& 42.9\\
Gemma3-27B & 1.60 & 0.88 & 1.51 & 1.44 & 1.63 & 1.22 & 0.69 & 0.97 & 1.16 & 1.29 & 88.3 \\
Gemini 2.5 Flash & 1.20 & 0.77 & 0.84 & 1.81 & 1.71 & 1.28 & 0.88 & 1.84 & 1.30 & 1.29 & 98.1 \\

\midrule
\multicolumn{12}{l}{\textit{Finetuned on MapTrace (ours)}} \\
Gemma3-27B* & 1.36\textsubscript{\textcolor{green}{-0.24}} & 1.03\textsubscript{\textcolor{red}{+0.15}} & 
1.02\textsubscript{\textcolor{green}{-0.49}} & 1.18\textsubscript{\textcolor{green}{-0.26}} & 
1.30\textsubscript{\textcolor{green}{-0.33}} & 1.00\textsubscript{\textcolor{green}{-0.22}} & 
0.83\textsubscript{\textcolor{red}{+0.14}} & 1.53\textsubscript{\textcolor{red}{+0.56}} & 
1.13\textsubscript{\textcolor{green}{-0.03}} & 1.13\textsubscript{\textcolor{green}{-0.16}} & 94.7\textsubscript{\textcolor{green}{+6.4}} \\
Gemini 2.5 Flash* & 1.14\textsubscript{\textcolor{green}{-0.06}} & 0.74\textsubscript{\textcolor{green}{-0.03}} & 
0.71\textsubscript{\textcolor{green}{-0.13}} & 0.93\textsubscript{\textcolor{green}{-0.88}} & 
1.04\textsubscript{\textcolor{green}{-0.67}} & 0.64\textsubscript{\textcolor{green}{-0.64}} & 
0.62\textsubscript{\textcolor{green}{-0.26}} & 0.93\textsubscript{\textcolor{green}{-0.91}} & 
0.90\textsubscript{\textcolor{green}{-0.40}} & \textbf{0.87}\textsubscript{\textcolor{green}{-0.42}} & \textbf{98.3}\textsubscript{\textcolor{green}{+0.2}} \\
\bottomrule
\end{tabular}}
\vspace{-.2cm}
\caption{
Performance on MapBench. Results are reported using Normalized Dynamic Time Warping (NDTW; lower is better) and Success Rate (SR; higher is better). 
Asterisks (*) denote models finetuned on our synthetic dataset.
}
\vspace{-.5cm}
\label{tab:maptrace_results}
\end{table*}

\begin{table}[t]
\centering
\resizebox{\linewidth}{!}{ 
\begin{tabular}{l l c c c}
\toprule
 & Setting & Items & Accuracy & False Positive Rate \\
\midrule
\multirow{2}{*}{}
  & Path Critic & 120 & 76\% & 8\% \\
  & Mask Critic & 200 & 83\% & 9\% \\
\bottomrule
\end{tabular}
}
\vspace{-.4cm}
\caption{Critic audits.}
\label{tab:critic_error_synth_quality}
\vspace{-0.6cm}
\end{table}

\vspace{-.2cm}
\section{MapTrace: Synthetic Data Generation}
\label{sec:data_generation}
\vspace{-.2cm}

Since collecting large-scale, pixel-accurate map annotations is prohibitively expensive, we design a fully synthetic pipeline to generate both maps and path traces automatically. Figure~\ref{fig:data_generation} illustrates the process, which consists of four stages: (i) map image generation, (ii) path mask extraction, (iii) pixel-graph construction, and (iv) final path generation. Each stage produces intermediate visualizations that allow MLLMs to act as ``critics,'' filtering low-quality samples and ensuring dataset quality.

\noindent\textbf{Map Image Generation.} We begin by prompting an LLM with diversified descriptions of map types (e.g., zoos, museums, amusement parks, shopping malls) and aspect ratios (details in Appendix~\ref{app:map_types}). These descriptions are instantiated into stylized map images using a text-to-image model. This step produces visually diverse maps that resemble real-world wayfinding designs, adhering to the language model's layout commonsense, while remaining annotation-free.

\noindent\textbf{Path Mask Generation. } To identify traversable regions, we extract dominant colors from the generated image and convert them into candidate binary masks. We use $k$-means clustering in RGB pixels of the map to segment the image into $k$ color clusters (details in Appendix~\ref{app:path_masks}). Each cluster corresponds to a candidate path region, which is evaluated by a \textbf{Mask Critic} LLM. Color-masks judged as valid are merged to form the final traversable path mask. This step ensures that accurate path regions are passed to downstream stages.

\noindent\textbf{Pixel-Graph Construction. } The validated path mask is then converted into a graph representation to support robust shortest-path computation. We discretize the mask into non-overlapping blocks and define nodes as blocks containing traversable pixels. Edges are added between nearby nodes, weighted by both spatial distance and pixel density (see Appendix~\ref{app:pixel_graph} for the formal definition). This representation captures the connectivity of the map while remaining computationally efficient.

\noindent\textbf{Final Path Generation. } Finally, we sample start and end points on the pixel-graph that are separated by at least a minimum distance $d_{\text{min}}$. The shortest path between these points is computed using Dijkstra’s algorithm and post-processed with the Ramer–Douglas–Peucker algorithm~\citep{saalfeld1999topologically} to reduce the number of path coordinates while preserving shape. Candidate paths are evaluated by a \textbf{Path Critic} MLLM, which verifies that they lie on valid traversable regions and avoid major obstacles. The paths that survive these checks form our synthetic dataset of 23k path annotations across 4k maps. 
\vspace{-.3cm}
\section{Experimental Setup}
\vspace{-.3cm}
\noindent\textbf{Data.} For training, we generate 23k synthetic paths across 4k distinct map images, with 20k used for training and 3k for validation. We evaluate on MapBench~\citep{xing2025can}, which contains 96 real-world maps (1,573 queries) collected from public sources, with human-provided path annotations.  \\
\noindent\textbf{Models.} Mask-Method is a non-learning baseline that extracts path regions directly from color-segmented masks. We also evaluate Gemma3-27B~\citep{team2025gemma} (open-weight) and Gemini-2.5-Flash~\citep{comanici2025gemini} (proprietary), each used as base models and further fine-tuned on our MapTrace dataset. Finally, to illustrate the difficulty of the task, we evaluate additional MLLMs such as  GLM4.5V~\cite{vteam2025glm45vglm41vthinkingversatilemultimodal}, QwenVL-2.5 72B~\cite{qwen2.5-VL}. \\
\noindent\textbf{Implementation Details.} Full hyperparameters and implementation configurations/infrastructure are provided in Appendix~\ref{appendix:impl_details}.
\vspace{-.3cm}
\section{Results}
\vspace{-.3cm}
\noindent\textbf{Finetuning on MapTrace.} We finetune models on our dataset using normalized coordinates at four decimal places of precision, with ablations on coordinate representation reported in Appendix~\ref{sec:appendix_ablations}. Table~\ref{tab:maptrace_results} shows that finetuning enables models to surpass strong baselines such as Gemini 2.5 Pro. Gemini 2.5 Flash improves markedly after finetuning, while Gemma3-27B also surpasses Pro in SR despite being weaker in its base form. These results highlight that synthetic supervision consistently boosts robustness and substantially reduces path alignment error. Notably, some apparent regressions in NDTW arise because models now solve a larger fraction of harder queries: higher success rates expand coverage to challenging cases that pretrained models failed to attempt, which inflates average distance despite overall gains.  MapTrace provides a scalable way to instill spatial reasoning that pretraining alone does not yield.\\
\noindent\textbf{Critics Audits.}  While LLM-based critics inevitably introduce some judgment noise, we empirically verify that their errors do not meaningfully degrade dataset quality. We conduct targeted manual audits of both critics: We manually review 120 \textbf{path critic} decisions across 56 randomly sampled maps. The critic achieves 76\% accuracy with only an 8\% false-positive rate (i.e., truly invalid paths marked as GOOD). The dominant error modes are (i) misclassifying background regions as traversable when background colors resemble path colors, and (ii) overlooking thin valid paths embedded within larger open regions. We manually inspect 200 \textbf{mask critic} judgments over 20 maps. The critic achieves 83\% accuracy with a 9\% false-positive rate. Typical mistakes include (i) background pixels being included due to color similarity with accurate paths, (ii) minor non-path elements such as text being absorbed into an otherwise correct mask, and (iii) thin but valid paths occasionally labeled as invalid.\\
\noindent\textbf{Human Inspection.} To evaluate the quality of the synthetic data we sampled 200 data points for human inspection across three categories: \texttt{High Quality}, \texttt{Acceptable}, \texttt{Low Quality}.  Annotators judged 58\% of samples as high quality and only 18\% as low quality, with moderate agreement (Cohen’s $\kappa=0.36$). Details are in Appendix~\ref{sec:appendix_human}.
\vspace{-.2cm}
\section{Conclusion}
\vspace{-.2cm}
In this work, we introduce a new method for evaluating map understanding in MLLMs: models must generate an exact sequence of coordinates that forms a valid path between two points while remaining within traversable regions. We show that this task is challenging even for state-of-the-art systems. To overcome the difficulty of collecting and annotating real maps at scale, we build an automated pipeline that uses text-to-image models to generate synthetic maps and leverages MLLMs as critics to select high-quality samples. This approach yields a dataset of 23k paths across 4k synthetic maps. Finetuning on this dataset leads to substantial performance gains on real-world maps sourced from the web.

\vspace{-.2cm}
\section{Limitations and Future Work}
\vspace{-.2cm}
While MapTrace advances the study of fine-grained spatial reasoning in MLLMs, several limitations remain.  
First, our dataset is fully synthetic. Although we designed the maps to mimic real-world layouts via LLM prompting for a consistent description and varied categories, they may still fail to capture the full diversity, and stylistic variation found in commercial wayfinding maps. While our evaluation on real maps shows its effectiveness and applicability, there is still a risk of domain shift when applying models trained on MapTrace to unseen maps in practice. Future work could explore semi-synthetic or real annotated maps to improve generalization, assuming permissive licenses for the real map artifacts.

Second, our evaluation is restricted to coordinate-level path tracing. While this directly probes spatial reasoning, it does not encompass downstream tasks such as multimodal navigation instructions, embodied agents, or reasoning over dynamic environments. Future work will extend MapTrace to use natural language instructions or even embodied interactions in real or simulated environments. 

Third, our pipeline relies on LLM-based ``critics'' for filtering masks and paths. These automated judgments are cost-effective, but imperfect as shown by our manual audits in Table \ref{tab:critic_error_synth_quality}. Future work could incorporate human-in-the-loop validation to increase robustness.  

Finally, although finetuning improves both robustness and geometric fidelity, the improvements are not perfectly reflected in NDTW scores. In particular, the substantial increase in success rate means that models are now producing paths for many harder queries they previously failed on, which raises the average NDTW despite better overall performance. Future work could explore hybrid evaluation strategies the jointly measure success rate and path alignment. 

Looking forward, we see MapTrace as a foundation for developing models that generalize from synthetic supervision to real-world navigation tasks, opening the door to future benchmarks and applications in robotics, AR/VR, and assistive/accessibility technologies.

\noindent\textbf{Note on AI Assistants. } During the preparation of this work, we made limited use of AI assistance for tasks such as proofreading, text rephrasing for clarity, and drafting figure/table captions. All conceptual contributions, experimental design, data generation, analysis, and interpretation of results were conducted by the authors.

\bibliography{custom}

\appendix

\appendix
\section{Synthetic Data Generation Details}
\label{sec:appendix_datagen}

This appendix provides details of the assets, prompts, and algorithms used in our synthetic pipeline.

\subsection{Map Categories}
\label{app:map_types}
We defined a diverse set of categories to generate representative map environments:
{\tiny
\begin{verbatim}
MAP_CATEGORIES = [
    "zoo",
    "urban",
    "botanical garden",
    "museum",
    "amusement park",
    "national park",
    "hospital",
    "hotel",
    "airport",
    "shopping mall",
    "restaurant",
    "campus"
]
\end{verbatim}
}

\subsection{Prompt Templates}
We employed LLMs both to generate map descriptions and to filter low-quality masks and paths. Below we include the key prompts used in the pipeline.

\paragraph{Map Description Generator.}
{\tiny
\begin{verbatim}
Create a short description of a top-view 2D pathfinding map of a {{ map_type }}. 
The description should be under 100 words and specify map style, landmarks, 
and key routes. The map should not have a title.
\end{verbatim}
}

\paragraph{Mask Critic.}
{\tiny
\begin{verbatim}
Your goal is to evaluate a segmentation mask's accuracy in identifying paths 
based on a source image.

You will be given two images:
* `[Image 1: Source]` is the original map image.
* `[Image 2: Mask]` is the segmentation mask where path/target areas should be 
  marked in white.

**Evaluation Criteria:**

* **Target Areas (Considered 'Correct'):** Any type of map paths: paved 
  sidewalks, marked crosswalks, pedestrian-only paths, public plazas, indoor 
  walkways etc.
* **Non-Target Areas (Considered 'Incorrect'):** Vehicle lanes of roads, grass, 
  dirt, buildings, cars, and any other non-pedestrian surface.
* If the majority of the image is white, it is likely of poor quality.

**Analysis Task:**

Instead of a simple Yes/No, please provide the following structured analysis of 
`[Image 2: Mask]`:

1.  **Composition Analysis:** Estimate the composition of the total white area in 
    the mask. Break it down by the type of ground it covers, with approximate 
    percentages.
    * *Example: The mask covers approximately 70% sidewalks, 25% vehicle 
      roadway, and 5% grass.*

2.  **Major Errors:** List the most significant `Non-Target Areas` that were 
    incorrectly included in the mask.

3.  **Final Judgment:** Based on your analysis, provide a final one-word 
    judgment on the mask's precision: `Good`, `Fair`, or `Poor`.
    * `Good`: The mask is almost entirely composed of Target Areas (>60%).
    * `Fair`: The mask contains a mix of Target and Non-Target areas.
    * `Poor`: The mask is predominantly composed of Non-Target areas (<40% correct).
\end{verbatim}
}

\paragraph{Path Critic.}
{\tiny
\begin{verbatim}
You will be shown an image of a map that has a magenta line drawn on it. 
Your task is to evaluate if this line represents a valid, traversable route.

Evaluation Criteria:

Boundary Check: The entire magenta line must be located within the main 
geographical area of the map. The line is invalid if any portion of it touches 
or crosses onto the image background, the map's title, the legend, scale bars, 
or any other non-map elements.

Traversability Check: The magenta line must follow a path that is considered 
traversable on the map (e.g., a road, a street, a marked trail, a walkway). 
The line is invalid if it cuts across buildings, solid land without a path, 
bodies of water (unless on a bridge or ferry route), or other impassable 
obstacles.

Your Response:

First, provide a step-by-step analysis of the line based on the two criteria 
above. In the final line of your response, state your final judgment as a 
single word: GOOD or BAD.
\end{verbatim}
}

\subsection{Color Clustering for Path Masks}
\label{app:path_masks}
Given an RGB input map image $\mathcal{I} \in \mathbb{R}^{W \times H \times 3}$, we treat its $N = W \times H$ pixels as RGB 3-element vectors $\{p_i\}$. We apply $k$-means clustering to partition pixels into $k$ color clusters:
\[
\underset{S}{\arg\min} \sum_{j=1}^k \sum_{p_i \in S_j} \|p_i - \mu_j\|^2,
\]
where $\mu_j$ is the centroid of cluster $S_j$. Centroids define dominant colors; cluster masks are binarized, visualized, and passed to the Mask Critic for filtering.

\subsection{Pixel-Graph Construction}
\label{app:pixel_graph}
The validated mask is transformed into a weighted undirected graph $G=(V,E,w)$:
\begin{itemize}
    \item \textbf{Nodes:} Non-overlapping $b \times b$ blocks containing $\geq$1 traversable pixel.  
    \item \textbf{Edges:} Connect nodes whose centers $c_i, c_j$ satisfy $\|c_i - c_j\|_2 \leq \delta_{\max}$.  
    \item \textbf{Weights:} Edge cost is distance penalized by node densities $\rho_i$:  
    \[
    w(v_i,v_j) = \|c_i - c_j\|_2 \cdot \big(1 + \lambda((1-\rho_i)+(1-\rho_j))\big).
    \]
\end{itemize}

\subsection{Path Sampling and Simplification}
We sample start and end nodes at least $d_{\min}$ apart, compute shortest paths with Dijkstra’s algorithm, and simplify them with the Ramer–Douglas–Peucker algorithm~\citep{saalfeld1999topologically}. Candidate paths are then filtered by the Path Critic.

\section{MapTrace Finetuning Ablations}
\label{sec:appendix_ablations}
\noindent\textbf{Path Trace Representation Choices} We experiment with different representation choices for path-tracing. While MLLMs are optimized for linguistic content, MapTrace requires the output of explicit coordinates. While existing works in object detection~\citep{yin2025rod,wu2024dettoolchain,tang2025visual} have explored coordinate representation in MLLMs, no explicit research has explored the best representation for an output that consists of a sequence of coordinates. 

\begin{table}[t]
    \centering
    \scriptsize
    \begin{tabularx}{.5\textwidth}{llrr}
    \toprule
    \multicolumn{2}{c}{\textbf{Coord. Representation}}  & \textbf{NDTW} $\downarrow$ & \textbf{SR} $\uparrow$\\
    \midrule
    \multirow{2}{*}{Absolute} & Base & 1.25 & 99.24\% \\
    & FT & 1.05 & 99.24\% \\
    \midrule
   \multirow{2}{*}{Normalized} &  Base &1.31 & 99.05\%\\
  & FT & 0.87 & 99.55\%\\
    \midrule
    \multirow{2}{*}{Delta} & Base & 1.01 & 96.74\%\\
     & FT & 1.24 & 98.03\%\\
    \bottomrule
\end{tabularx}
\vspace{-.3cm}
\caption{Coordinate representation performance on MapBench (\texttt{gemini-2.5-flash}).}
\label{tab:coord_representation}
\end{table}

\begin{table}[t]
    \centering
    \scriptsize
        \begin{tabularx}{.5\textwidth}{llrr}
    \toprule
    \multicolumn{2}{c}{\textbf{Coord. Precision}}  & \textbf{NDTW} $\downarrow$ & \textbf{SR} $\uparrow$\\
    \midrule
    \multirow{1}{*}{Full} &  & 0.91 & 99.55\% \\
    \midrule
   \multirow{1}{*}{4 Decimals} &   &0.87 & 99.55\%\\
    \midrule
    \multirow{1}{*}{3 Decimals} &  & 0.99 & 99.49\%\\
        \midrule
    \multirow{2}{*}{2 Decimals} &  & 1.14 & 99.55\%\\

    \bottomrule
\end{tabularx}
\vspace{-.3cm}
\captionof{table}{Coordinate precision performance on MapBench (\texttt{gemini-2.5-flash}).}
\label{tab:coord_precision}
\end{table}

We explore three natural options: \textbf{absolute coordinates}, \textbf{normalized coordinates}, and \textbf{delta changes} from previous positions. In Table \ref{tab:coord_representation} we show that normalized coordinates perform best on the base models likely due to pretraining on object detection with normalized coordinates. 

\noindent\textbf{Coordinate Precision:} Given that normalized coordinates perform the best, the next natural question for data representation is what is the optimal precision level for the coordinate system. We experiment with full precision, precision of 4,3 and 2 decimal places.
 The results in Table \ref{tab:coord_precision} show that under 4 points of precision there are significant losses in performance.
 
 \section{MapTrace Dataset}
 \label{app:data_examples}
\begin{figure*}[h]
\vspace{-0cm} \centering
\includegraphics[width=\linewidth]{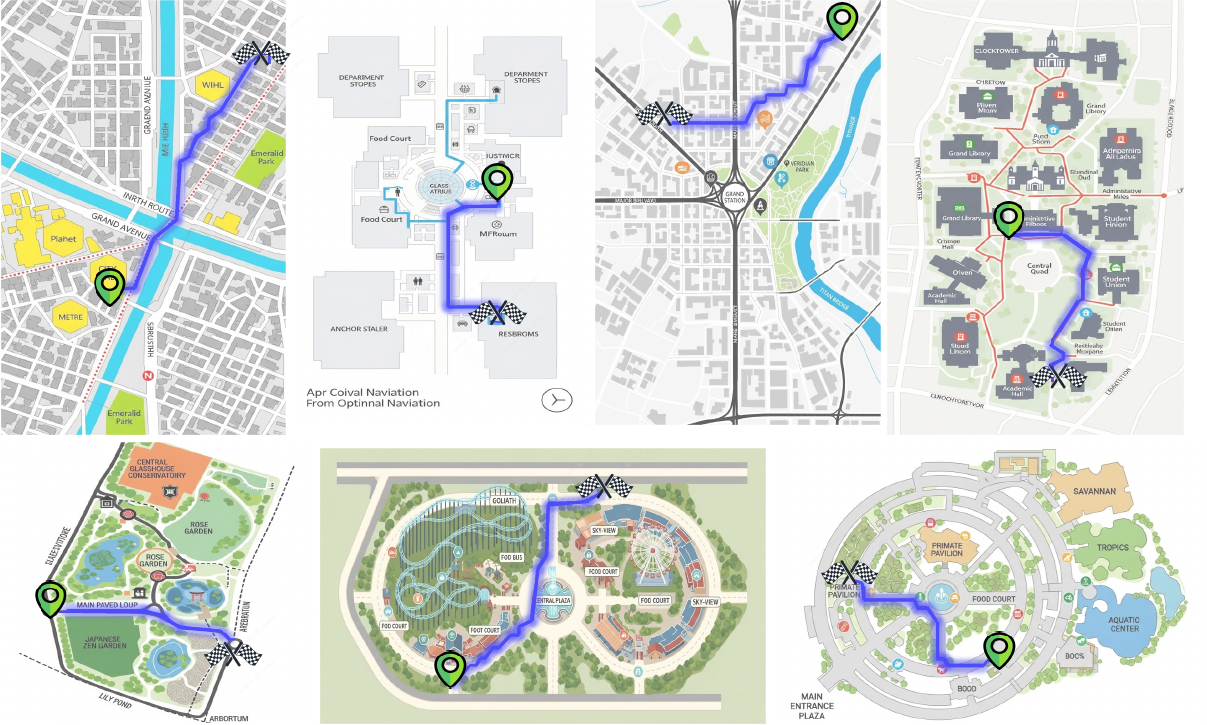}
\vspace{-.7cm}
    \caption{Qualitative examples of MapTrace paths on diverse map types.}
\vspace{-.5cm}
    \label{fig:qualitative_extra}
\end{figure*}

 Figure \ref{fig:qualitative_extra} shows a sample of examples from the generated dataset. 
\subsection{Data Distribution}
\begin{figure}[h]
    \centering 
    \includegraphics[width=.7\linewidth]{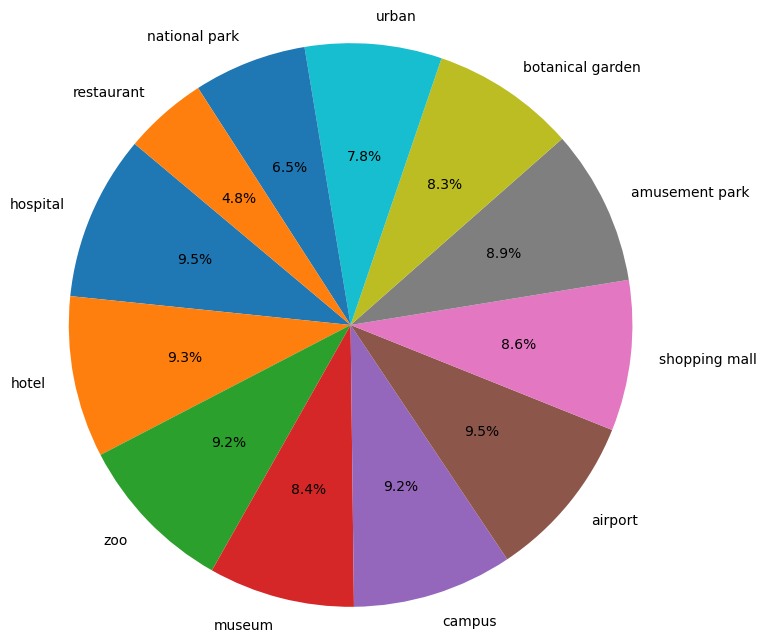}
    \caption{Distribution of map types in the synthetic MapTrace dataset.}
    \label{fig:data_dist}
\end{figure}

Figure~\ref{fig:data_dist} shows the distribution of map types in our synthetic dataset, spanning diverse environments such as zoos, museums, campuses, malls, and national parks.
\section{Implementation Details}
\label{appendix:impl_details}

\subsection{Infrastructure and Data Synthesis}
We use Vertex-AI\footnote{\tiny\href{https://cloud.google.com/vertex-ai}{https://cloud.google.com/vertex-ai}} to serve and fine-tune Gemini models, and Simula~\citep{davidson2025orchestrating} to orchestrate data synthesis. Imagen-4.0~\citep{saharia2022photorealistic} is used as the text-to-image generator, with aspect ratios sampled uniformly from \{1:1, 3:4, 4:3, 16:9, 9:16\}. Mask generation uses an RGB tolerance of 25. For the pixel-graph, we set \texttt{max\_distance=4}, \texttt{block\_size=4}, \texttt{density\_penalty=50}, and a minimum start–end distance of 200 pixels. We perform a single run for each experiment and report the result in Table \ref{tab:maptrace_results}.

\subsection{Gemini Models}
Gemini-2.5-Pro~\citep{comanici2025gemini} is used as the underlying MLLM for the description generator, mask critic, and path critic modules. For fine-tuning, we run 5 epochs with LoRA (adapter rank 8), batch size 16, and learning rate 0.001.  

\subsection{Gemma Models}
For Gemma3-27B~\citep{team2025gemma}, we fine-tune with LoRA adapters (rank 8), batch size 16, and learning rate 5e-5. Training runs for 5 epochs with mixed precision (bfloat16). Finetuning is completed on 8A100 40GB GPUs and is completed in 10 hours. 

\subsection{Mask Baseline}
The \textit{Mask-Method} baseline directly uses dominant color segmentation to produce path masks. Shortest paths are then computed within these masks using Dijkstra’s algorithm. This provides a non-learning comparison that highlights the benefit of fine-tuning MLLMs.

\subsection{Data and Licenses}
MapBench~\citep{xing2025can} is released under the Apache License 2.0, which permits use, modification, and distribution for both academic and commercial purposes.  
We will release the \textbf{MapTrace synthetic dataset} under a \textbf{research-only, non-commercial license}. 

\section{Human Inspection Instructions}
\label{sec:appendix_human}

The following instructions were provided to human evaluators for judging the final quality of the generated map traces. Two human annotators with advanced technical degrees were recruited on a volunteer basis to conduct this investigation. Annotators were shown an image with a path plotted and asked to categorize its quality based on these criteria.

\begin{itemize}
    \item[]\textbf{High Quality:} The trace strictly adheres to designated, traversable paths and follows an efficient, logical route. The only acceptable reason for deviating from a path is to traverse from a start/end position to the traversable areas.
    
    \item[]\textbf{Acceptable Quality:} The trace generally follows traversable paths but is inefficient or includes minor deviations into plausible off-path areas (e.g., cutting across a lawn).
    
    \item[]\textbf{Low Quality:} The trace is implausible, crossing clearly non-traversable barriers (e.g., buildings, bodies of water without a bridge) or consistently deviating from the designated paths.
\end{itemize}

\begin{figure}[t]
    \centering
    \includegraphics[width=\linewidth]{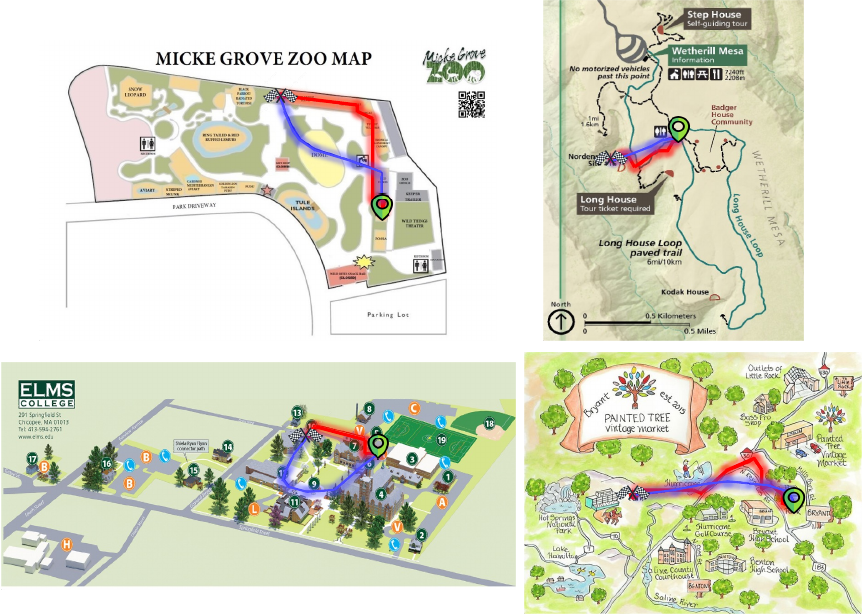}
    \caption{Qualitative examples comparing the fine-tuned Gemini-2.5-Flash (\textcolor{red}{red}) to the base model (\textcolor{blue}{blue}). The fine-tuned model adheres more closely to the intended routes and avoids non-traversable regions.}
    \label{fig:mapbench_examples}
\end{figure}

\section{Qualitative Examples}
Figure~\ref{fig:mapbench_examples} shows that the fine-tuned model follows the intended paths rather than trivially connecting endpoints or crossing non-traversable obstacles. We inspected the 50 highest-NDTW predictions from the best finetuned model (\texttt{gemini-2.5-flash}) and categorized the dominant error types. The distribution is as follows: \textbf{Invalid Detours into Non-Traversable Regions (26\%):} The model deviates into clearly non-path areas such as buildings or fenced-off regions, often due to misinterpreting visually similar textures as walkable space (e.g. large green areas). \textbf{Overly Long but Still Valid Paths (14\%):} The model remains entirely on valid traversable regions but chooses longer routes rather than the shortest path.
\textbf{Early Off-Path Drift Followed by Recovery (14\%):} Initial segments drift slightly outside the walkable areas before the model realigns with the correct path.
\textbf{Correct Path Start but Failure in Long-Horizon Planning (14\%):} The model begins on path regions but fails to maintain global directionality, often committing to a branch that leads to a dead-end and then compensating by moving off-path in an attempt to reach the goal.

\end{document}